\documentclass[lettersize,journal]{IEEEtran}
\usepackage{amsmath,amsfonts}
\usepackage{array}
\usepackage[caption=false,font=normalsize,labelfont=sf,textfont=sf]{subfig}
\usepackage{textcomp}
\usepackage{stfloats}
\usepackage{url}
\usepackage{verbatim}
\usepackage{graphicx}
\usepackage{booktabs}
\usepackage{multirow}
\usepackage{makecell}
\usepackage{tabularx}
\usepackage{placeins}
\usepackage{pifont}
\usepackage{tabularx}
\usepackage{makecell}
\usepackage{array}
\usepackage{makecell}
\usepackage{algorithm}
\usepackage{algpseudocode}
\usepackage{booktabs}
\usepackage{array}
\usepackage{makecell}
\newcolumntype{C}[1]{>{\centering\arraybackslash}m{#1}}

\algrenewcommand\algorithmicrequire{\textbf{Input:}}
\algrenewcommand\algorithmicindent{0.8em}

\newcommand{\cmark}{\ding{51}}
\usepackage{cite}
\usepackage{etoolbox}

\makeatletter
\patchcmd{\IEEEbiographynophoto}
  {\vskip 4\baselineskip plus 1fil minus 0\baselineskip}
  {\vskip 2.3\baselineskip plus 4.0\baselineskip}
  {}{}
\makeatother

\begin{document}

\title{VitaTouch: Property-Aware Vision–Tactile–Language Model for Robotic Quality Inspection in Manufacturing}

\author{Junyi~Zong, Qingxuan~Jia, Meixian~Shi, Tong~Li, Jiayuan~Li, Zihang~Lv, Gang~Chen, and Fang~Deng,~\IEEEmembership{Fellow,~IEEE}%
\thanks{Junyi Zong, Qingxuan Jia, Meixian Shi, Tong Li, Zihang Lv, and Gang Chen are with the School of Intelligent Engineering and Automation, Beijing University of Posts and Telecommunications, Beijing 100876, China.}%
\thanks{Junyi Zong and Jiayuan Li are also with Zhongguancun Academy, Beijing 100094, China.}%
\thanks{Jiayuan Li is with the School of Automation, Beijing Institute of Technology, Beijing 100081, China.}%
\thanks{Fang Deng is with the School of AI, Beijing Institute of Technology, Beijing 100081, China, and also with the Beijing Key Laboratory of Lightweight Intelligent System, Beijing Institute of Technology, Beijing 100081, China.}%
\thanks{Corresponding authors: Tong Li and Fang Deng (e-mail: tli@bupt.edu.cn; dengfang@bit.edu.cn).}%
}

\maketitle

\begin{abstract}
Quality inspection in smart manufacturing requires identifying intrinsic material and surface properties beyond visible geometry, yet vision-only methods remain vulnerable to occlusion and reflection. We propose VitaTouch, a property-aware vision--tactile--language model for material-property inference and natural-language attribute description. VitaTouch uses modality-specific encoders and a dual Q-Former to extract language-relevant visual and tactile features, which are compressed into prefix tokens for a large language model. We align each modality with text and explicitly couple vision and touch through contrastive learning. We also construct VitaSet, a multimodal dataset with 186 objects, 52k images, and 5.1k human-verified instruction--answer pairs. VitaTouch achieves the best performance on HCT and the overall TVL benchmark, while remaining competitive on SSVTP. On VitaSet, it reaches 88.89\% hardness accuracy, 75.13\% roughness accuracy, and 54.81\% descriptor recall; the material-description task further achieves a peak semantic similarity of 0.9009. With LoRA-based fine-tuning, VitaTouch attains 100.0\%, 96.0\%, and 92.0\% accuracy for 2-, 3-, and 5-category defect recognition, respectively, and delivers 94.0\% closed-loop recognition accuracy and 94.0\% end-to-end sorting success in 100 laboratory robotic trials. More details are available at the project page: https://vitatouch.github.io/.
\end{abstract}


\section{Introduction}

Quality inspection is a core component of intelligent manufacturing, directly influencing product reliability, production efficiency, and lifecycle cost. Rather than being limited to end-of-line verification, it spans incoming material inspection, in-process monitoring, and pre-delivery defect assessment. In zero-defect manufacturing, automated in-line inspection enables real-time defect detection and process adaptation for full-process quality control \cite{azamfirei2023automation}. Robot-based intelligent inspection further supports defect identification, classification, and closed-loop parameter adjustment to suppress defect propagation before early degradation evolves into structural failure \cite{lu2023deeplearning}. From the perspective of Industry 4.0, inspection strategies should be continuously updated using production data and quality-cost tradeoff analysis to balance conformance, efficiency, and lifecycle decisions \cite{reis2024inspection}.

Despite these advances, industrial inspection is still largely dominated by vision-based pipelines. Under controlled conditions and with sufficient training data, modern machine vision and deep models achieve strong performance on industrial benchmarks. Large-scale datasets and surveys have demonstrated the effectiveness of deep industrial anomaly detection (IAD) at both image and pixel levels \cite{liu2024iad_survey,bergmann2019mvtec}, and systematic reviews have confirmed the maturity and broad adoption of vision-based surface defect inspection \cite{ma2024surface_review}. However, when inspection is framed as physical understanding rather than appearance recognition, purely visual approaches exhibit inherent limitations.

Vision provides only a partial basis for physical quality assessment. While it captures geometric and photometric appearance, many manufacturing-critical properties, such as \textbf{hardness}, \textbf{roughness}, and \textbf{material composition}, are not directly observable. Its reliability is further compromised by practical shop-floor conditions, including \textbf{occlusions}, visually inaccessible regions, \textbf{uneven illumination}, and \textbf{specular reflections} on metallic or glossy surfaces, which degrade appearance cues and hinder robust deployment \cite{ma2024surface_review,liu2024iad_survey}. Reliable inspection therefore requires sensing modalities that can directly access the physical attributes of manufactured surfaces.

Recent advances in tactile sensing, particularly in robotic perception and manipulation, offer such complementary physical evidence \cite{li2024tactilegrasping,xin2025vbtsreview}. Vision-based tactile sensors (VBTS) record contact-induced deformation through RGB tactile readouts, thereby encoding local surface and material properties while being less susceptible to illumination artifacts than conventional vision \cite{xin2025vbtsreview}. Recent progress in visuo-tactile representation learning has further shown the potential of transferable tactile foundations across sensors and data regimes \cite{feng2025anytouch}, motivating property-centric inspection grounded in contact mechanics for quality assessment and early defect detection.

At the same time, foundation models and large language models (LLMs) have reshaped multimodal learning by linking low-level sensory observations with high-level semantic concepts. Vision-language pretraining, exemplified by CLIP, enables scalable image-language alignment for open-vocabulary recognition and flexible prompting \cite{radford2021clip}, while multimodal LLMs such as BLIP-2, InstructBLIP, and LLaVA support instruction following and open-ended generation conditioned on sensory inputs \cite{li2023blip2,dai2023instructblip,liu2023llava}. For inspection, this shift opens the possibility of language-driven workflows in which operators specify criteria in natural language and models return open-form attribute descriptions and reasoning-oriented explanations.

Early studies have begun to integrate touch with language-grounded perception, including touch-vision-language (TVL) alignment on the TVL dataset \cite{fu2024tvl} and tactile-language models for object property reasoning such as Octopi \cite{yu2024octopi}. However, these efforts focus on general robotic perception and everyday objects rather than the stricter requirements of industrial quality inspection. Recent benchmarks further show that existing multimodal LLMs remain inadequate for industrial anomaly and defect understanding \cite{jiang2024mmad}. A reusable tactile-grounded, property-centric foundation for physical understanding in smart manufacturing inspection is therefore still lacking.

Practical deployment introduces additional constraints. Defect annotations are often scarce and fragmented across product variants, and representative defective samples are difficult to collect because of low defect incidence and long-tail variability. Meanwhile, industrial PCs and edge devices impose strict latency and reliability requirements under limited GPU memory and compute budgets. These conditions make full-model finetuning impractical, especially for large multimodal models. Effective solutions must therefore support few-shot adaptation with minimal labeled data through resource-efficient updates while preserving real-time inference. Existing methods, however, lack property-aware multimodal grounding, resulting in fragile transfer when appearance changes or defects are primarily manifested through physical property variations.

In this work, we identify \textbf{multimodal physical-property understanding} as the missing foundation for reliable industrial quality inspection. We propose \textbf{VitaTouch}, a \textbf{vision-tactile-language (VTL) property-aware multimodal model} for embodied robotic quality inspection. VitaTouch learns shared grounding among visual observations, tactile evidence, and natural language queries to support multi-property inference and language-form material/surface understanding. It follows a property-centric pretraining-to-deployment paradigm: establishing cross-modal alignment across vision, touch, and language; learning reusable material and surface property representations; and enabling efficient downstream defect inspection adaptation in few-shot, resource-constrained settings via parameter-efficient updates (e.g., low-rank adaptation). By treating defect inspection as a downstream task of general property-aware pretraining, VitaTouch better matches the multi-stage nature of industrial quality inspection and supports transfer from property understanding to practical defect recognition.

Our main contributions are summarized as follows:
\begin{itemize}
\item We construct and release a curated vision-tactile-language (VTL) dataset for robotic manipulation, built on our industrial subset and the GelSight subset of AnyTouch with a unified annotation schema, providing aligned visual, tactile, and language annotations for multimodal quality inspection.
\item We propose a property-centric training pipeline coupling cross-modal alignment with multi-property attribute learning in the VTL setting, establishing a reusable backbone for material and surface property understanding.
\item We introduce VitaTouch, a property-aware vision-tactile-language model for smart manufacturing multimodal quality inspection, targeting unified physical understanding rather than appearance-only detection.
\item We validate efficient few-shot defect recognition transfer under resource constraints via parameter-efficient adaptation, demonstrating that property-centric pretraining supports practical industrial inspection scenarios.
\end{itemize}

\begin{figure*}[t]
    \centering
    \includegraphics[width=\textwidth]{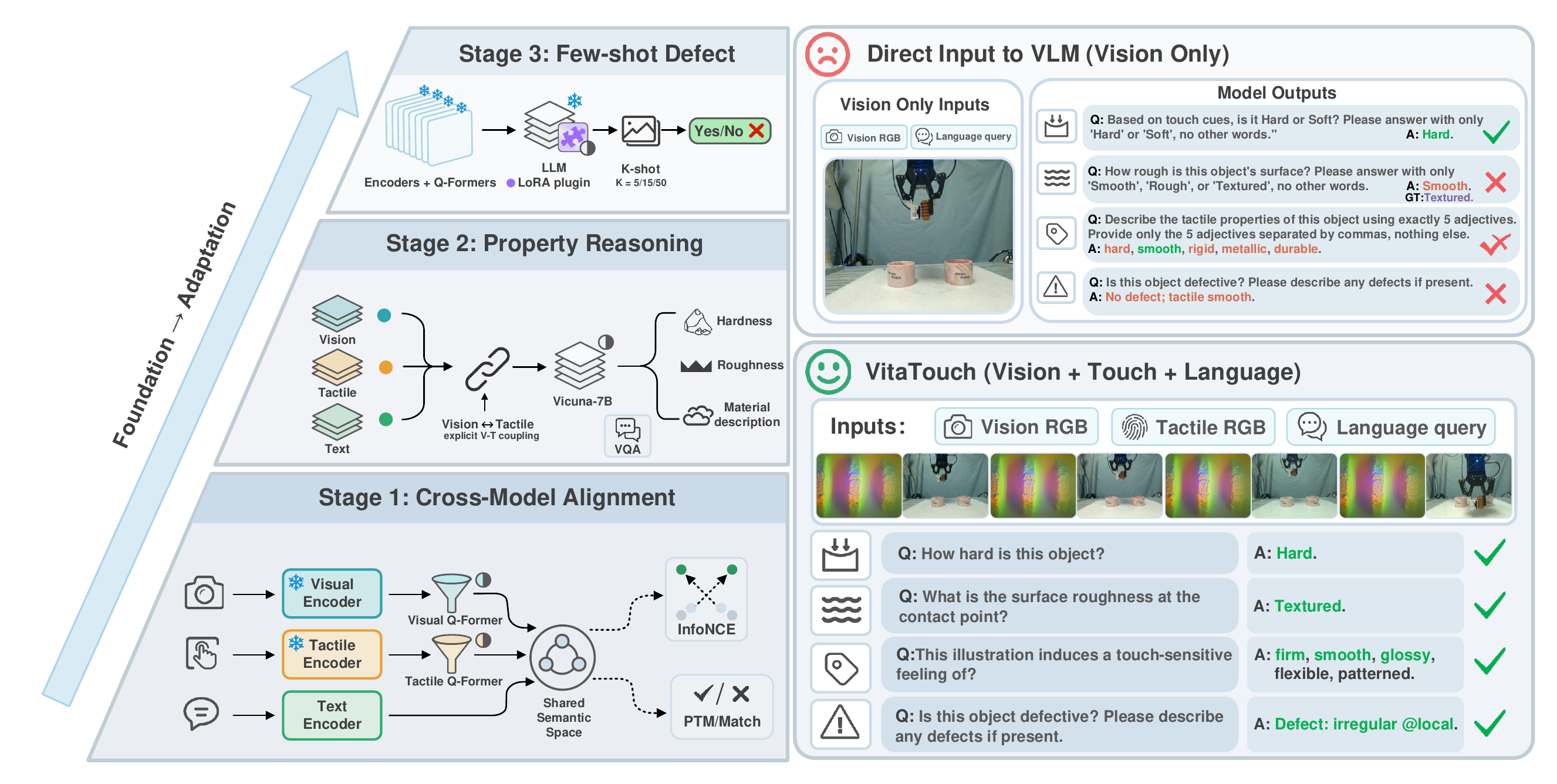}
    \caption{\textbf{Overview of VitaTouch.} Left: three-stage training pipeline. Stage~1 performs cross-modal alignment via dual Q-Formers with InfoNCE and PTM losses. Stage~2 builds a property-reasoning multimodal model with fused V--T tokens in frozen Vicuna-7B. LoRA-based defect adaptation is then conducted over progressively finer-grained defect label spaces using few-shot labeled samples per category. Right: tactile sensing complements vision for property reasoning and defect recognition under visually ambiguous conditions.}
    \label{fig:framework}
\end{figure*}

\section{Related Works}
\subsection{Vision-Based Defect Inspection with Deep Learning}
Vision-based inspection dominates industrial quality control, where deep learning has significantly advanced defect detection and localization. Though fully supervised models perform well in controlled settings with dense annotations, real-world applications are limited by the scarcity, diversity, and high annotation cost of defective samples, driving the development of industrial anomaly detection (IAD) methods trained primarily on normal data. Representative IAD directions include feature distribution modeling (e.g., PaDiM) \cite{defard2021padim}, discriminative reconstruction and embedding learning (e.g., DRAEM) \cite{zavrtanik2021draem}, and retrieval-based patch embedding matching (e.g., PatchCore) \cite{roth2022patchcore}. MVTec AD \cite{bergmann2019mvtec} serves as the standard benchmark, with VisA \cite{zou2022spd} broadening evaluation diversity and recent TMECH work further exploring multi-view unsupervised inspection for complex industrial components \cite{zhou2025mvagfl}, while recent surveys systematically review deep IAD and surface defect inspection advances \cite{liu2024iad_survey,ma2024surface_review}. Despite these progresses, purely visual pipelines remain vulnerable to occlusion, illumination variation, and specular reflection, with inherent limitations for defects tied to physical properties rather than stable visual appearance.

\subsection{From Closed-Set Recognition to Language-Driven Inspection with LLM/VLMs}
Recent advances in large language models (LLMs) and vision-language models (VLMs) are shifting industrial inspection from closed-set recognition to language-driven reasoning. Compared with traditional multi-class classifiers, LLM/VLM-based systems enable natural language instruction following, open-ended defect description beyond fixed taxonomies, and semantic reasoning for interactive operator-centric workflows. CLIP-style pretraining establishes scalable image-text alignment for promptable open-vocabulary recognition \cite{radford2021clip}, and multimodal LLMs (BLIP-2, InstructBLIP, LLaVA) further couple visual perception with powerful language decoders to support instruction tuning and generative multimodal responses \cite{li2023blip2,dai2023instructblip,liu2023llava}. However, industrial inspection imposes domain-specific requirements including fine-grained localization, long-tail defect adaptation, and safety-critical precision. State-of-the-art multimodal LLMs still fail to meet industrial anomaly and defect understanding requirements \cite{jiang2024mmad}, motivating methods that integrate language capabilities while complying with manufacturing constraints.

\subsection{VLM/LLM for Defect and Anomaly Detection: Methods and Benchmarks}
A growing body of work adapts CLIP-style VLMs for anomaly and defect detection, leveraging language to improve cross-category generalization. A core challenge is that generic VLMs are biased toward object semantics rather than normality-abnormality distinction, making naive prompting ineffective for anomaly localization. WinCLIP addresses this via window- and patch-level feature aggregation with compositional prompt ensembling, achieving strong zero-/few-shot performance on MVTec AD and VisA \cite{jeong2023winclip}.
Prompt learning reduces reliance on hand-crafted prompts: PromptAD enables one-class prompt learning with only normal samples for few-shot detection \cite{li2024promptad}, while AnomalyCLIP learns object-agnostic prompts for cross-domain zero-shot anomaly detection \cite{zhou2024anomalyclip}. Context-conditioned prompting and cross-modal interaction have also been explored, with VCP-CLIP injecting visual context into prompts to avoid product-specific design \cite{qu2024vcpclip} and recent TMECH work extending prompt-based inspection to multimodal RGB--3D settings through cross-modal prompt learning and contrastive pretraining under missing-modality conditions \cite{jiang2025resilient}. Recent works combine multiple foundation models to improve localization: ClipSAM integrates CLIP-based anomaly localization with SAM mask refinement for sharper boundaries \cite{li2024clipsam}. Anomaly-aware alignment is also introduced to separate normal and abnormal semantics in CLIP features, with AA-CLIP using anomaly-aware text anchors and patch-level alignment while preserving generalization \cite{ma2025aaclip} and FE-CLIP incorporating frequency cues for improved zero-shot performance \cite{gong2025feclip}. Beyond CLIP-centric pipelines, AnomalyGPT realizes dialogic interaction and threshold-free anomaly reporting via anomaly-oriented prompting and synthetic supervision \cite{gu2023anomalygpt}. Evaluations on mainstream benchmarks consistently show that language significantly improves the scalability, flexibility, and interpretability of defect inspection systems \cite{bergmann2019mvtec,zou2022spd,jiang2024mmad}.

\subsection{Introducing Touch: Deep Visuo-Tactile Learning and LLM-Grounded Touch--Vision--Language}
While language-guided VLMs improve semantic flexibility, reliable inspection requires modalities capturing surface and material cues beyond visual appearance. Vision-based tactile sensors (VBTS), which output image-like tactile observations encoding micro-geometry and contact mechanics, are increasingly practical for industrial use \cite{xin2025vbtsreview}, with representative platforms including GelSight \cite{yuan2017gelsight} and the compact DIGIT sensor \cite{lambeta2020digit}. For tactile modeling, AnyTouch explores unified static-dynamic representations across multiple visuo-tactile sensors for transferable encoding \cite{feng2025anytouch}. In industrial settings, global vision combined with local tactile scanning has proven effective for detecting small defects on large components, with coarse-to-fine strategies mitigating visual coverage limitations and blind spots \cite{agarwal2023roboticdefect}.
Tactile research is also evolving from visuo-tactile fusion to LLM-grounded tactile reasoning with natural language. The TVL dataset, a key milestone, provides large-scale vision-touch pairs with language annotations to enable tactile encoder alignment with vision-language spaces and tactile-conditioned text generation \cite{fu2024tvl}. Octopi further demonstrates tactile-language property reasoning by combining tactile representation learning with VLMs, leveraging intermediate hardness and roughness predictions to improve downstream performance \cite{yu2024octopi}. However, these efforts mainly target general robotics and daily object understanding, and industrial inspection still lacks a reusable, property-centric touch-vision-language foundation tailored to manufacturing defect taxonomies, process variability, and deployment constraints.

\subsection{Few-Shot Adaptation and Parameter-Efficient Tuning for Industrial Deployment}
Industrial quality inspection differs from general multimodal reasoning in three core aspects. First, defect data is scarce: failure modes are rare, product-specific, and require domain expertise for annotation, incurring high labeling costs. Second, production environments are dynamic, with frequent new variants and process drift requiring rapid adaptation without full retraining. Third, deployment faces strict computational constraints, with edge devices such as industrial PCs imposing tight latency and memory limits that hinder direct application of large multimodal models.
Two learning paradigms address these challenges. For data scarcity, few-shot learning and meta-learning enable rapid adaptation from limited samples, with representative methods including MAML \cite{finn2017maml} and prototypical networks \cite{snell2017protonet}. For resource constraints, parameter-efficient tuning adapts foundation models by updating only lightweight modules while freezing the backbone, with LoRA significantly reducing adaptation costs via trainable low-rank updates \cite{hu2021lora}. Nevertheless, most existing few-shot anomaly inspection methods remain vision-centric, lacking property-aware multimodal grounding. For defects better characterized by physical property deviations than visual patterns, the absence of tactile evidence and explicit property understanding reduces model robustness under distribution shift, motivating multimodal, property-centric pretraining for robust few-shot transfer in manufacturing.

\section{Dataset and Training \& Evaluation Suite}
\label{sec:dataset_suite}

This section introduces \textbf{VitaSet}, a vision--tactile--language (VTL) dataset and accompanying training/evaluation suite for industrial quality inspection. Existing visuo-tactile datasets mainly emphasize daily objects and general robotic perception, offering limited support for industrial inspection scenarios and task-specific supervision. To address this limitation, we construct VitaSet by combining an in-house industrial robotic manipulation subset with the GelSight-only subset of AnyTouch under a unified annotation schema. The resulting dataset supports both property-centric pretraining and inspection-oriented evaluation for VitaTouch.

\subsection{Physical Property Selection}
\label{subsec:property_selection}

We focus on three physical properties that are central to industrial inspection. They are well-defined, closely related to common nonconformities, and suitable for multimodal perception.

\noindent\textbf{Hardness.} Hardness characterizes resistance to indentation and deformation under applied pressure. It is an important inspection attribute in manufacturing, and deviations may indicate material mismatch or process drift.

\noindent\textbf{Roughness.} Roughness describes surface micro-geometry and directly affects friction, wear resistance, sealing performance, and coating adhesion.

\noindent\textbf{Material and Surface Characteristics.} Material and surface characteristics are closely associated with mechanical and tribological behavior and are therefore relevant to incoming inspection, contamination prevention, and production-line sorting. In VitaSet, this property is formulated as a language description task, in which the model describes material- and surface-related attributes from multimodal observations.

\subsection{Dataset Construction and Annotation}
\label{subsec:data_collection}

We define the released dataset as
\begin{equation}
\mathcal{D}=\{(\boldsymbol{X}_i, T_i, Y_i)\}_{i=1}^{N}, \qquad
\boldsymbol{X}_i=(\boldsymbol{X}^{\mathrm{v}}_i, \boldsymbol{X}^{\mathrm{t}}_i),
\end{equation}
where $\boldsymbol{X}_i^{\mathrm{v}}$ denotes the RGB image, $\boldsymbol{X}_i^{\mathrm{t}}$ the GelSight tactile image, $T_i$ the instruction/query, and $Y_i$ the corresponding text-format answer.

VitaSet integrates two complementary sources. The first is an in-house industrial robotic manipulation dataset collected with an RGB camera and a GelSight sensor during repeated grasp-and-contact interactions that emulate inspection operations. The second is the GelSight-only subset of AnyTouch, included to increase object and contact diversity while preserving tactile-modality consistency across sources.

\begin{figure*}[!t]
\centering
\includegraphics[width=\textwidth]{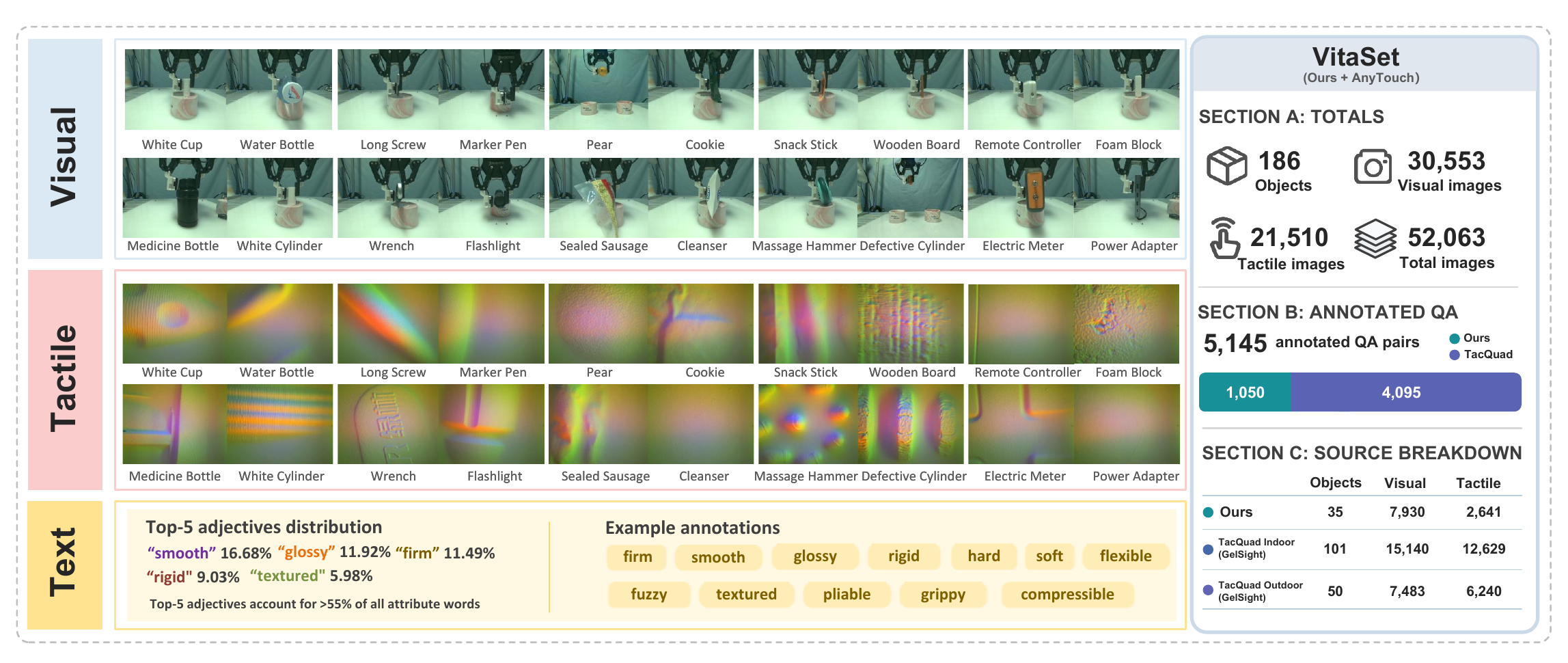}
\caption{\textbf{VitaSet overview (Ours + AnyTouch GelSight-only).} Aligned RGB observations and paired GelSight tactile readings across objects, with controlled-vocabulary annotations and dataset statistics under a unified schema.}
\label{fig:dataset_overview}
\end{figure*}

In total, VitaSet contains \textbf{186 objects}, \textbf{30,553} RGB images, \textbf{21,510} GelSight tactile images, and \textbf{5,145} annotated instruction--answer pairs. The smaller number of tactile images is due to manual removal of invalid or low-quality contact frames. An overview is shown in Fig.~\ref{fig:dataset_overview}.

To ensure annotation quality and consistency, candidate QA pairs are first generated using \textbf{GPT-4o} with template-constrained prompting. Classification tasks are restricted to predefined label sets, whereas description tasks follow a fixed output format with a controlled vocabulary. All generated annotations are then fully reviewed and corrected by human annotators for both the in-house and AnyTouch-derived subsets. This full-coverage verification over all \textbf{186 objects} yields a unified and auditable annotation space across the entire dataset.

\subsection{Training and Evaluation Suite}
\label{subsec:task_suite}

We define four tasks aligned with industrial quality inspection, all cast in the form $(\boldsymbol{X}, T)\rightarrow Y$, where $\boldsymbol{X}=(\boldsymbol{X}^{\mathrm{v}}, \boldsymbol{X}^{\mathrm{t}})$ denotes the multimodal input.

\noindent\textbf{Task 1: Binary Hardness Classification.} The model predicts one label from $\{\texttt{Hard},\,\texttt{Soft}\}$. Evaluation uses classification accuracy.

\noindent\textbf{Task 2: 3-Way Roughness Classification.} The model predicts one label from $\{\texttt{Smooth},\,\texttt{Textured},\,\texttt{Rough}\}$. Evaluation uses classification accuracy.

\noindent\textbf{Task 3: Material-Property Description.} The model generates a short textual description of material and surface properties from multimodal observations. Evaluation includes descriptor recall based on lexical matching against human-annotated reference descriptors and a supplementary semantic similarity score.

\noindent\textbf{Task 4: Defect Recognition.} The model outputs a text-form label from a predefined defect taxonomy. We consider progressively finer 2-, 3-, and 5-category settings, corresponding to defect presence detection, defect type recognition, and joint defect type/location recognition. Evaluation uses classification accuracy.

Overall, the suite covers both property-oriented physical understanding and inspection-oriented defect reporting, supporting foundation-model pretraining and downstream adaptation for industrial inspection.

\section{Method}
\label{sec:method}

VitaTouch is a vision--tactile--language model for property reasoning and few-shot defect inspection. As shown in Fig.~\ref{fig:framework}, we train VitaTouch with a progressive three-stage protocol: (i) cross-modal alignment to establish a shared semantic interface across vision, touch, and language; (ii) property-reasoning foundation learning by conditioning a frozen LLM on fused vision--tactile tokens; and (iii) parameter-efficient few-shot adaptation to defect recognition using LoRA. 

\subsection{Problem Formulation}
\label{subsec:problem}

Each sample consists of an RGB image, a tactile image, and a natural-language query,
\begin{equation}
\mathcal{X}_i = \big(\boldsymbol{X}^{\mathrm{v}}_i,\, \boldsymbol{X}^{\mathrm{t}}_i,\, T_i\big),
\end{equation}
where $\boldsymbol{X}^{\mathrm{v}}_i \in \mathbb{R}^{H_{\mathrm{v}} \times W_{\mathrm{v}} \times 3}$ denotes the visual observation, $\boldsymbol{X}^{\mathrm{t}}_i \in \mathbb{R}^{H_{\mathrm{t}} \times W_{\mathrm{t}} \times 3}$ denotes the tactile observation, and $T_i$ specifies either a property query or a defect query. Given $\mathcal{X}_i$, VitaTouch generates a textual response $Y_i$.

\subsection{Model Architecture}
\label{subsec:arch}

\begin{figure*}[t]
  \centering
  \includegraphics[width=\textwidth]{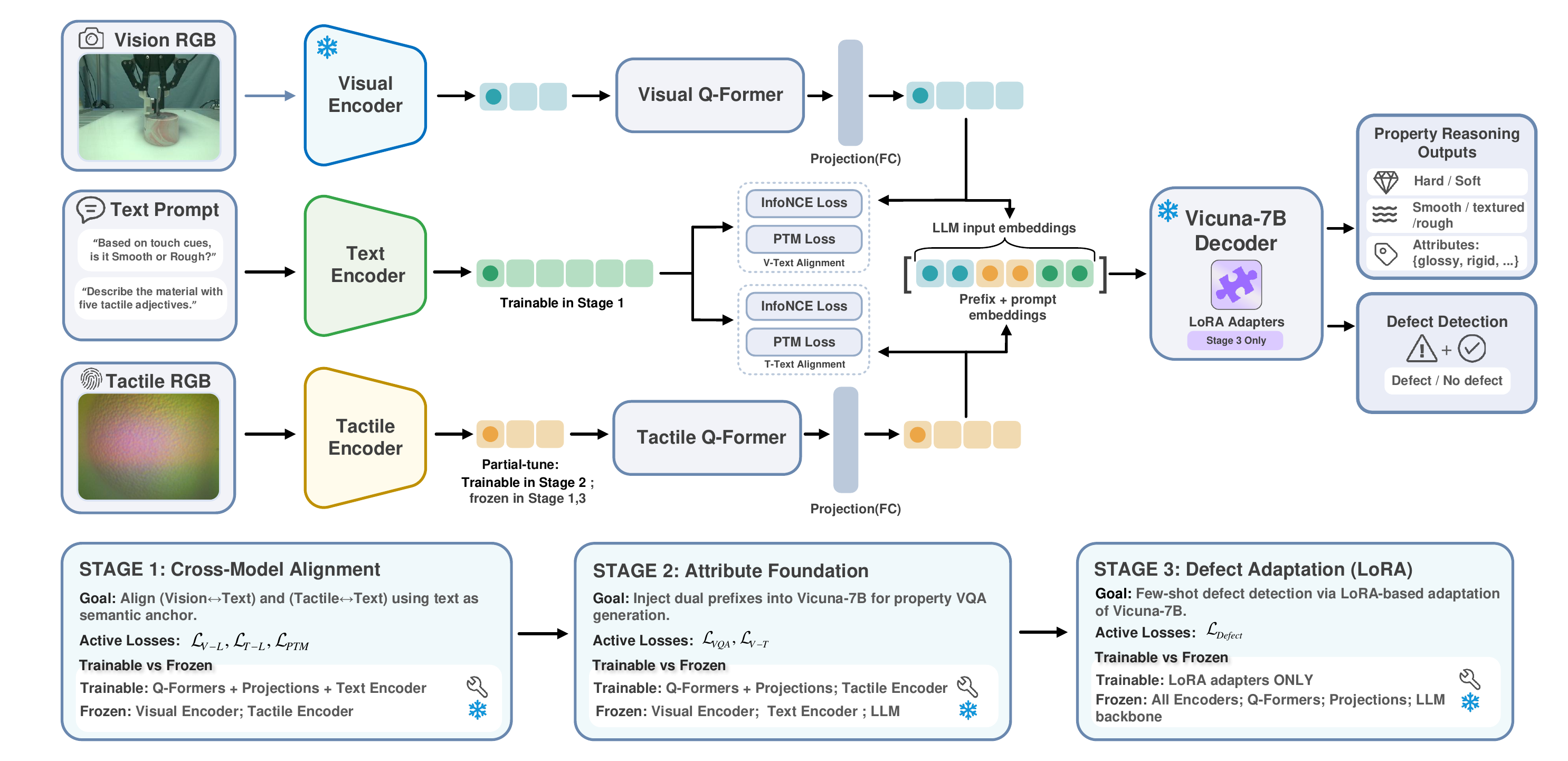}
  \caption{\textbf{VitaTouch model architecture.}
VitaTouch employs a \textbf{dual-branch vision-tactile design} with modality-specific encoders and Q-Formers. 
The Q-Formers distill learnable queries into \textbf{vision-tactile prefix tokens}, which are prepended to text embeddings and fed into a frozen Vicuna-7B decoder. 
Training proceeds in three stages: 
\textbf{Stage 1} aligns cross-modal embeddings via frozen encoders; 
\textbf{Stage 2} establishes the perception-to-language pathway for property reasoning; 
\textbf{Stage 3} adapts the model for few-shot defect recognition using LoRA while freezing the backbone.}
  \label{fig:vitamodel}
\end{figure*}

An overview of the proposed architecture is provided in Fig.~\ref{fig:vitamodel}. VitaTouch follows a dual-branch design with modality-specific encoders and Q-Formers. The Q-Formers distill compact, language-relevant tokens from vision and touch, which are then mapped into the LLM embedding space as prefix tokens.

\subsubsection{Perception Encoders}
\label{subsec:encoders}

Given $\boldsymbol{X}^{\mathrm{v}}_i$, the visual encoder produces a token sequence
\begin{equation}
\boldsymbol{E}^{\mathrm{v}}_i = f^{\mathrm{v}}_{\theta}\!\big(\boldsymbol{X}^{\mathrm{v}}_i\big) \in \mathbb{R}^{L_{\mathrm{v}} \times d_{\mathrm{v}}}.
\end{equation}
Given $\boldsymbol{X}^{\mathrm{t}}_i$, the tactile encoder produces
\begin{equation}
\boldsymbol{E}^{\mathrm{t}}_i = f^{\mathrm{t}}_{\theta}\!\big(\boldsymbol{X}^{\mathrm{t}}_i\big) \in \mathbb{R}^{L_{\mathrm{t}} \times d_{\mathrm{t}}}.
\end{equation}
The visual encoder is frozen throughout all stages. The tactile encoder is frozen in Stage~1 and fully fine-tuned in Stage~2, while it is kept frozen in Stage~3.

\subsubsection{Dual Q-Former}
\label{subsec:qformer}

For each modality $m \in \{\mathrm{v},\mathrm{t}\}$, we instantiate an independent Q-Former with $L_{\mathrm{q}}$ learnable query vectors,
\begin{equation}
\boldsymbol{Q}^{m}_0 \in \mathbb{R}^{L_{\mathrm{q}} \times d_{\mathrm{q}}}, 
\qquad L_{\mathrm{q}} = 32 .
\end{equation}
Given perception tokens $\boldsymbol{E}^{m}_i$, the Q-Former outputs
\begin{equation}
\boldsymbol{Q}^{m}_i = \mathrm{QF}^{m}\!\big(\boldsymbol{Q}^{m}_0,\, \boldsymbol{E}^{m}_i\big)
\in \mathbb{R}^{L_{\mathrm{q}} \times d_{\mathrm{q}}}.
\end{equation}
Each query token is projected into a shared embedding space and $\ell_2$-normalized:
\begin{equation}
\hat{\boldsymbol{q}}^{m}_{ik} =
\frac{\boldsymbol{W}^{m}_{\mathrm{p}}\, \boldsymbol{Q}^{m}_{ik}}  
{\big\|\boldsymbol{W}^{m}_{\mathrm{p}}\, \boldsymbol{Q}^{m}_{ik}\big\|_2},
\qquad
\hat{\boldsymbol{q}}^{m}_{ik} \in \mathbb{R}^{d},
\quad k = 1,\dots,L_{\mathrm{q}}.
\label{eq:qformer_proj}
\end{equation}

\subsubsection{Text Encoding and LLM Interface}
\label{subsec:text_llm}

Given query text $T_i$, a text encoder produces token states $\boldsymbol{Z}_i$. We project the \texttt{[CLS]} state into the shared embedding space:
\begin{equation}
\hat{\boldsymbol{t}}_i =
\frac{\boldsymbol{W}_{\mathrm{t}}\, \boldsymbol{Z}_{i,\texttt{[CLS]}}}
{\big\|\boldsymbol{W}_{\mathrm{t}}\, \boldsymbol{Z}_{i,\texttt{[CLS]}}\big\|_2}
\in \mathbb{R}^{d}.
\label{eq:text_proj}
\end{equation}

For Stage~2/3, query tokens are mapped into the LLM embedding space:
\begin{equation}
\tilde{\boldsymbol{Q}}^{\mathrm{v}}_i,\; \tilde{\boldsymbol{Q}}^{\mathrm{t}}_i \in \mathbb{R}^{L_{\mathrm{q}} \times d_{\mathrm{LLM}}},
\qquad
\tilde{\boldsymbol{Q}}^{\mathrm{vt}}_i
=
\big[\tilde{\boldsymbol{Q}}^{\mathrm{v}}_i;\, \tilde{\boldsymbol{Q}}^{\mathrm{t}}_i\big]
\in \mathbb{R}^{2L_{\mathrm{q}} \times d_{\mathrm{LLM}}}.
\end{equation}

\subsection{Stage~1: Cross-Modal Alignment Pretraining}
\label{subsec:stage1}

Stage~1 learns a shared semantic interface by aligning each modality with language. For a minibatch of size $N$, we pool the query set by averaging:
\begin{equation}
\hat{\boldsymbol{z}}^{m}_i
=
\frac{1}{L_{\mathrm{q}}}\sum_{k=1}^{L_{\mathrm{q}}}\hat{\boldsymbol{q}}^{m}_{ik}
\in \mathbb{R}^{d},
\qquad
m \in \{\mathrm{v},\mathrm{t}\}.
\label{eq:pooled_z}
\end{equation}

We apply symmetric bidirectional InfoNCE between modality $m$ and language, where $\tau$ denotes the temperature parameter:
\begin{equation}
\begin{split}
\mathcal{L}_{m\text{-}L}
= -\frac{1}{2N}\sum_{i=1}^{N} \Bigg[
&\log
\frac{\exp\!\big(\langle \hat{\boldsymbol{z}}^{m}_i,\hat{\boldsymbol{t}}_i\rangle/\tau\big)}
{\sum_{j=1}^{N}\exp\!\big(\langle \hat{\boldsymbol{z}}^{m}_i,\hat{\boldsymbol{t}}_j\rangle/\tau\big)}
\\
&\quad+
\log
\frac{\exp\!\big(\langle \hat{\boldsymbol{t}}_i,\hat{\boldsymbol{z}}^{m}_i\rangle/\tau\big)}
{\sum_{j=1}^{N}\exp\!\big(\langle \hat{\boldsymbol{t}}_i,\hat{\boldsymbol{z}}^{m}_j\rangle/\tau\big)}
\Bigg].
\end{split}
\label{eq:infonce}
\end{equation}

The overall contrastive loss is defined as the sum of the modality--language contrastive terms:
\begin{equation}
\mathcal{L}_{\mathrm{con}}
=
\mathcal{L}_{\mathrm{v}\text{-}L}
+
\mathcal{L}_{\mathrm{t}\text{-}L}.
\label{eq:contrastive_total}
\end{equation}

To encourage semantic compatibility beyond metric alignment, we use a perceptual-text matching (PTM) objective with hard negatives, where $M$ denotes the number of perceptual-text pairs, $y_{\ell}\in\{0,1\}$ is the ground-truth matching label for pair $\ell$, and $p_{\ell}$ is the predicted probability of a match.
\begin{equation}
\mathcal{L}_{\mathrm{match}}
=
-\frac{1}{M}\sum_{\ell=1}^{M}
\Big[
y_{\ell}\log p_{\ell} + (1-y_{\ell})\log(1-p_{\ell})
\Big],
\label{eq:matching_loss}
\end{equation}

The Stage~1 objective is
\begin{equation}
\mathcal{L}_{\mathrm{Stage1}} = \mathcal{L}_{\mathrm{con}} + \lambda_{\mathrm{match}}\mathcal{L}_{\mathrm{match}}.
\label{eq:stage1_loss}
\end{equation}
Here, $\lambda_{\mathrm{match}}$ controls the relative weight of the PTM objective.
\subsection{Stage~2: Property-Reasoning multimodal Model}
\label{subsec:stage2}

Stage~2 trains the model to answer property-related queries conditioned on fused vision--tactile prefixes. Given a property prompt $T^{\mathrm{prop}}_i$ and target sequence $Y_i=(y_{i,1},\dots,y_{i,L_i})$, where $L_i$ denotes the target length and $y_{i,<s}=(y_{i,1},\dots,y_{i,s-1})$ denotes the target prefix before step $s$, we minimize the autoregressive loss:
\begin{equation}
\mathcal{L}_{\mathrm{VQA}}
=
-\frac{1}{N}\sum_{i=1}^{N}\sum_{s=1}^{L_i}
\log p_{\theta}\!\big(y_{i,s}\mid y_{i,<s}, \boldsymbol{X}^{\mathrm{v}}_i, \boldsymbol{X}^{\mathrm{t}}_i, T^{\mathrm{prop}}_i\big).
\label{eq:VQA_loss}
\end{equation}

We further enforce explicit vision--tactile coupling with a symmetric InfoNCE loss computed between the pooled vision and tactile embeddings:
\begin{equation}
\begin{split}
\mathcal{L}_{\mathrm{v}\text{-}\mathrm{t}}
= -\frac{1}{2N}\sum_{i=1}^{N} \Bigg[
&\log
\frac{\exp\!\big(\langle \hat{\boldsymbol{z}}^{\mathrm{v}}_i,\hat{\boldsymbol{z}}^{\mathrm{t}}_i\rangle/\tau_{\mathrm{vt}}\big)}
{\sum_{j=1}^{N}\exp\!\big(\langle \hat{\boldsymbol{z}}^{\mathrm{v}}_i,\hat{\boldsymbol{z}}^{\mathrm{t}}_j\rangle/\tau_{\mathrm{vt}}\big)}
\\
&\quad+
\log
\frac{\exp\!\big(\langle \hat{\boldsymbol{z}}^{\mathrm{t}}_i,\hat{\boldsymbol{z}}^{\mathrm{v}}_i\rangle/\tau_{\mathrm{vt}}\big)}
{\sum_{j=1}^{N}\exp\!\big(\langle \hat{\boldsymbol{z}}^{\mathrm{t}}_i,\hat{\boldsymbol{z}}^{\mathrm{v}}_j\rangle/\tau_{\mathrm{vt}}\big)}
\Bigg].
\end{split}
\label{eq:v_t_coupling}
\end{equation}
where $\tau_{\mathrm{vt}}$ is a temperature hyperparameter.

The Stage~2 objective is
\begin{equation}
\mathcal{L}_{\mathrm{Stage2}} = \mathcal{L}_{\mathrm{VQA}} + \lambda_{\mathrm{VT}}\mathcal{L}_{\mathrm{v}\text{-}\mathrm{t}}.
\label{eq:stage2_loss}
\end{equation}

\subsection{Stage~3: LoRA-Based Defect Adaptation}
\label{subsec:stage3}

Stage~3 adapts the Stage~2 multimodal model to \textbf{defect recognition} under data-limited settings. To enable parameter-efficient transfer, we freeze the perception encoders, Q-Formers, and modality-to-LLM projection layers, and insert LoRA modules into the attention projections of the Vicuna decoder. Specifically, the adapted weight is written as
\begin{equation}
\boldsymbol{W}' = \boldsymbol{W} + \frac{\alpha}{r}\boldsymbol{BA},
\label{eq:lora}
\end{equation}
where $r$ is the LoRA rank and $\alpha$ is the scaling factor.

Given a defect-related prompt $T_i^{\mathrm{def}}$ and a target answer sequence $Y_i=(y_{i,1},\dots,y_{i,L_i})$, we optimize the standard autoregressive cross-entropy loss:
\begin{equation}
\mathcal{L}_{\mathrm{defect}}
=
-\frac{1}{N}\sum_{i=1}^{N}\sum_{s=1}^{L_i}
\log p_{\theta}\!\big(y_{i,s}\mid y_{i,<s}, \boldsymbol{X}^{\mathrm{v}}_i, \boldsymbol{X}^{\mathrm{t}}_i, T_i^{\mathrm{def}}\big),
\label{eq:defect_loss_lm}
\end{equation}
where $L_i$ denotes the token length of the target answer sequence $Y_i$, and
\(
y_{i,<s}=(y_{i,1},\dots,y_{i,s-1})
\)
denotes the target prefix before step $s$. In our setting, the target answer sequence typically corresponds to a defect-category label or \texttt{Normal} in text form.

At inference time, the model generates a textual response conditioned on the multimodal input and the defect prompt:
\begin{equation}
\hat{Y}_i \sim p_{\theta}\!\big(\cdot \mid \boldsymbol{X}^{\mathrm{v}}_i, \boldsymbol{X}^{\mathrm{t}}_i, T_i^{\mathrm{def}}\big).
\label{eq:defect_generation}
\end{equation}
The generated response is then mapped to the predefined coarse-to-fine defect taxonomy for evaluation. In this way, Stage~3 preserves the language-form prediction interface of VitaTouch while enabling efficient downstream adaptation for defect recognition.

\section{Experiments}
\label{sec:exp}

We evaluate VitaTouch on both public and manufacturing-oriented benchmarks. Experiments include comparison on the TVL benchmark, multi-task evaluation on VitaSet, ablation analysis, few-shot defect recognition via LoRA adaptation, qualitative case studies, and closed-loop robotic deployment.

\subsection{Datasets and Benchmarks}
\label{subsec:datasets}

\paragraph{TVL benchmark}
We evaluate VitaTouch on the public Touch--Vision--Language (TVL) benchmark~\cite{fu2024tvl}. Following the official protocol, we use the original benchmark setup and data split without introducing additional external data. Results are reported on the Self-Supervised Visuo-Tactile Pretraining (SSVTP) subset, the Human Collected Tactile (HCT) subset, and the overall TVL benchmark for direct comparison with prior methods.

\paragraph{VitaSet}
We further evaluate VitaTouch on VitaSet, our synchronized RGB--GelSight dataset for manufacturing-oriented physical understanding. Experiments cover hardness classification, roughness recognition, and material-property description, consistent with the property-reasoning setting introduced in Sec.~\ref{sec:dataset_suite}.

\subsection{Implementation Details}
\label{subsec:impl}

VitaTouch is implemented in PyTorch. We use EVA-CLIP-G as the visual encoder, an AnyTouch encoder with a ViT-L backbone for tactile inputs, and Vicuna-7B as the language backbone.

\paragraph{Stage~1: Cross-modal alignment}
Stage~1 trains the modality-specific Q-Formers and projection layers while keeping the perception encoders fixed. Optimization uses AdamW with a linear-warmup cosine schedule, a peak learning rate of $1\times10^{-4}$, a warmup learning rate of $1\times10^{-6}$, 500 warmup steps, and weight decay of 0.01. Automatic mixed precision (AMP) is used throughout training.

\paragraph{Stage~2: Property-reasoning foundation learning}
Stage~2 follows the stage-wise training strategy described in Sec.~\ref{sec:method}. We use AdamW with the same learning-rate schedule and weight decay as in Stage~1, and set the maximum input/output length to 512 tokens.

\paragraph{Stage~3: LoRA-based defect adaptation}
Stage~3 adapts the Stage~2 model to defect recognition by updating LoRA parameters only. Unless otherwise specified, the LoRA configuration uses rank $r=16$, scaling factor $\alpha=32$, and dropout rate 0.1.

\paragraph{Training hardware}
Stage~1 and Stage~2 are trained on an 8$\times$NVIDIA A800 GPU cluster, taking approximately 2 and 10 hours, respectively. Stage~3 is completed within 1 hour on a single RTX 5090 GPU.

\subsection{Main Results}
\label{subsec:results}

\subsubsection{Results on the TVL Benchmark}
\label{subsubsec:tvl_results}

We first evaluate VitaTouch on the public Touch--Vision--Language (TVL) benchmark~\cite{fu2024tvl}. For this benchmark, we strictly follow the official protocol, using only the original TVL dataset and its standard split for both training and evaluation, without introducing any external data or additional annotations. As shown in Table~\ref{tab:tvl_results_reproduced}, VitaTouch achieves the best performance on \textbf{HCT} and the overall \textbf{TVL} benchmark, while remaining competitive on \textbf{SSVTP}. These results demonstrate the effectiveness of VitaTouch under a standardized public benchmark and enable a fair comparison with prior methods.

\newcommand{\mname}[1]{\mbox{#1}}

\begin{table}[!t]
\centering
\scriptsize
\setlength{\tabcolsep}{3.2pt}
\renewcommand{\arraystretch}{1.05}
\caption{Comparison on the TVL benchmark. VitaTouch achieves the best performance on HCT and TVL, and remains competitive on SSVTP.}
\label{tab:tvl_results_reproduced}

\resizebox{\columnwidth}{!}{%
\begin{tabular}{@{}lcccccc@{}}
\toprule
\multicolumn{1}{c}{\textbf{Model}} &
\multicolumn{3}{c}{\textbf{Modalities}} &
\multicolumn{3}{c}{\textbf{Score}} \\
\cmidrule(lr){2-4}\cmidrule(l){5-7}
& \textbf{Vision}
& \textbf{Tactile}
& \textbf{Language}
& \textbf{SSVTP}
& \textbf{HCT}
& \textbf{TVL} \\
\midrule
LLaVA-1.5 7B             & \cmark & -      & \cmark & 3.64 & 3.55 & 3.56 \\
LLaVA-1.5 13B            & \cmark & -      & \cmark & 3.55 & 3.63 & 3.62 \\
ViP-LLaVA 7B             & \cmark & -      & \cmark & 2.72 & 3.44 & 3.36 \\
ViP-LLaVA 13B            & \cmark & -      & \cmark & 4.10 & 3.76 & 3.80 \\
LLaMA-Adapter            & \cmark & -      & \cmark & 2.56 & 3.08 & 3.02 \\
BLIP-2 Opt-6.7b          & \cmark & -      & \cmark & 2.02 & 2.72 & 2.64 \\
InstructBLIP 7B          & \cmark & -      & \cmark & 1.40 & 1.30 & 1.31 \\
InstructBLIP 13B         & \cmark & -      & \cmark & 1.44 & 1.21 & 1.24 \\
GPT-4V                   & \cmark & -      & \cmark & 5.02 & 4.42 & 4.49 \\
\midrule
SSVTP-LLaMA              & \cmark & \cmark & -      & 2.58 & 3.67 & 3.54 \\
\midrule
TVL-LLaMA (ViT-Tiny)     & \cmark & \cmark & \cmark & 6.09 & 4.79 & 4.94 \\
TVL-LLaMA (ViT-Small)    & \cmark & \cmark & \cmark & 5.81 & 4.77 & 4.89 \\
TVL-LLaMA (ViT-Base)     & \cmark & \cmark & \cmark & \textbf{6.16} & 4.89 & 5.03 \\
\midrule
VitaTouch (Ours)         & \cmark & \cmark & \cmark & 5.83 & \textbf{6.10} & \textbf{6.09} \\
\bottomrule
\end{tabular}%
}
\end{table}

\subsubsection{Property Understanding on VitaSet}
\label{subsubsec:selfbuilt_results}

We next evaluate VitaTouch on VitaSet for three property-related tasks: hardness classification, roughness classification, and material descriptor prediction. Hardness and roughness are evaluated by accuracy. For the material task, we report both strict descriptor recall and a supplementary semantic-similarity score.

Let $G_i$ denote the set of five human-annotated reference descriptors for sample $i$, and let $P_i$ denote the set of valid predicted descriptors after duplicate removal and invalid-word filtering. Descriptor recall is defined as
\begin{equation}
R_{\mathrm{desc}}=\frac{1}{N}\sum_{i=1}^{N}\frac{|P_i \cap G_i|}{|G_i|},
\end{equation}
where $N$ is the number of validation samples. To account for semantically similar but lexically different outputs, we further compute the mean cosine similarity between the predicted and ground-truth descriptor texts using normalized sentence embeddings extracted from Qwen3-Embedding-0.6B\cite{zhang2025qwen3embedding}:
\begin{equation}
\mathbf{z}_i^{\mathrm{pred}}=\frac{E_{\phi}(y_i^{\mathrm{pred}})}{\|E_{\phi}(y_i^{\mathrm{pred}})\|_2}, \qquad
\mathbf{z}_i^{\mathrm{gt}}=\frac{E_{\phi}(y_i^{\mathrm{gt}})}{\|E_{\phi}(y_i^{\mathrm{gt}})\|_2},
\end{equation}
\begin{equation}
S_{\mathrm{sem}}=\frac{1}{N}\sum_{i=1}^{N}(\mathbf{z}_i^{\mathrm{pred}})^{\top}\mathbf{z}_i^{\mathrm{gt}}.
\end{equation}

\begin{figure*}[!t]
  \centering
  \includegraphics[width=0.96\textwidth]{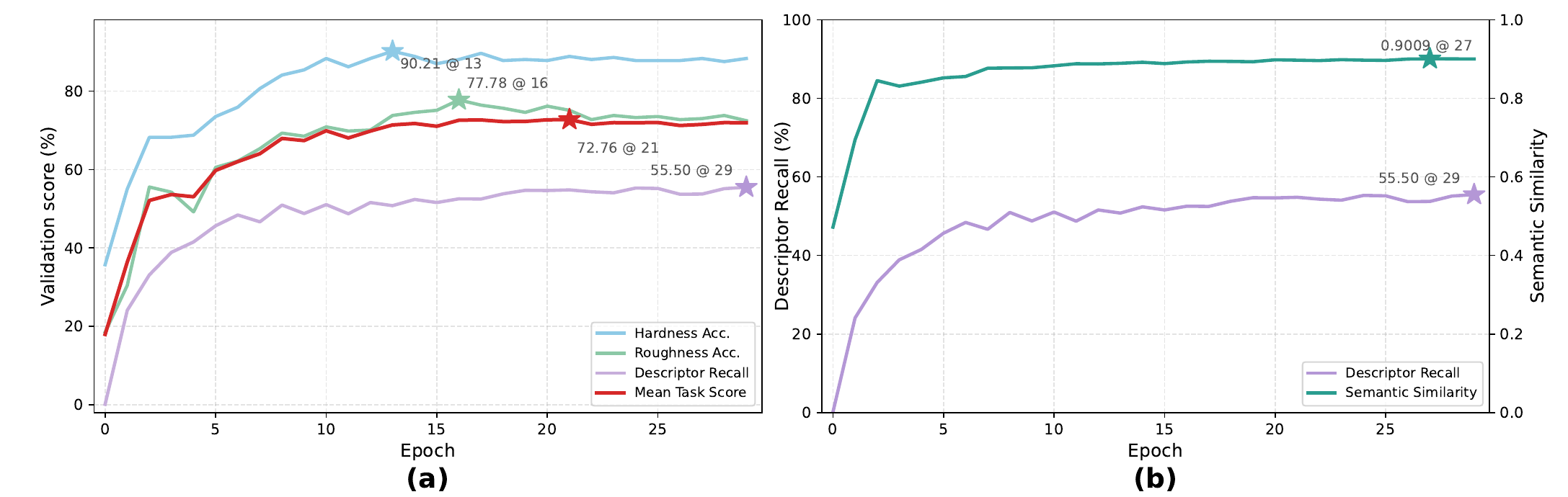}
  \caption{VitaSet validation performance of VitaTouch across training epochs. (a) Multi-task validation trends for hardness accuracy, roughness accuracy, descriptor recall, and mean task score. (b) Comparison of strict exact-match descriptor recall and semantic similarity for the material-property descriptor task. Stars ($\star$) mark the best epoch for each metric.}
  \label{fig:selfbuilt_alltasks_curve}
\end{figure*}

Unless otherwise stated, all reported results are evaluated on the held-out test set, while model checkpoint selection is performed on the validation set according to the highest mean task score. As shown in Fig.~\ref{fig:selfbuilt_alltasks_curve}, the selected checkpoint is epoch 21, achieving 88.89\% hardness accuracy, 75.13\% roughness accuracy, and 54.81\% descriptor recall, with a mean task score of 72.95\%. Different tasks peak at different epochs, but the selected checkpoint provides the best overall multi-task balance.

Fig.~\ref{fig:selfbuilt_alltasks_curve}(b) further shows that the semantic similarity remains high even when exact lexical matching is not achieved. In particular, the material-property descriptor task reaches a semantic similarity of 0.9009 at epoch 27, indicating that VitaTouch produces semantically aligned material descriptions beyond strict word-level overlap.

\subsubsection{Ablation Study}
\label{subsubsec:ablation}

\begin{figure*}[!t]
  \centering
  \includegraphics[width=0.92\textwidth]{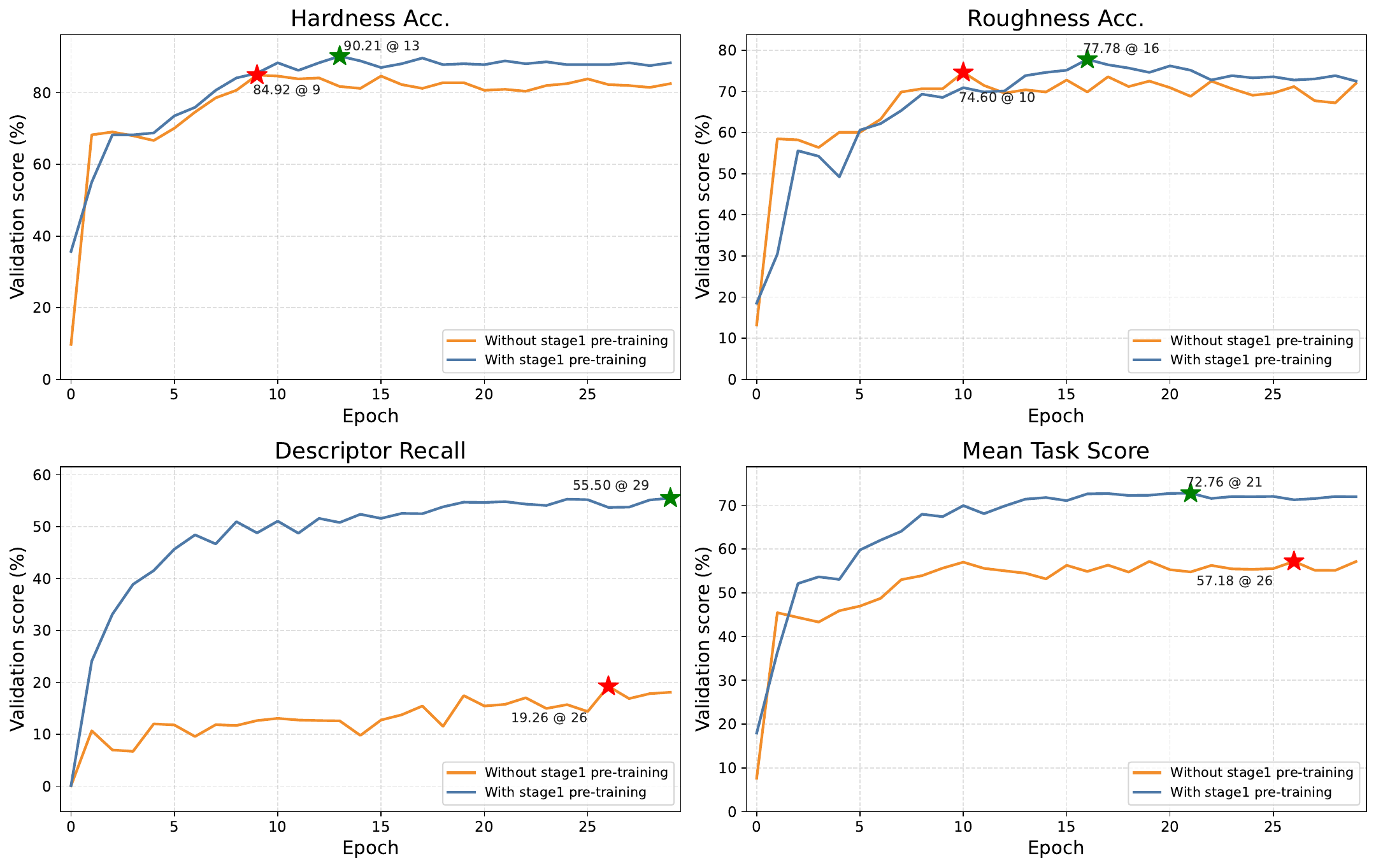}
  \caption{Ablation results on the VitaSet dataset across tasks. Each variant removes one key stage from the full model, demonstrating the necessity of explicit alignment and multimodal fusion for robust multi-task property learning.}
  \label{fig:selfbuilt_ablation_all4}
\end{figure*}

Table~\ref{tab:selfbuilt_summary} and Fig.~\ref{fig:selfbuilt_ablation_all4} report the ablation results on VitaSet. The full VitaTouch model achieves the best overall performance, reaching 88.89\% hardness accuracy, 75.13\% roughness accuracy, and 54.81\% descriptor recall, corresponding to a mean task score of 72.95\%.

Both unimodal baselines perform substantially worse than the full model, especially on the descriptor task, indicating that neither appearance cues nor localized tactile observations alone are sufficient for robust material-oriented language prediction. Removing Stage~1 alignment also leads to a clear degradation across all tasks, with descriptor recall dropping from 54.81\% to 17.46\%. This result confirms that explicit vision--tactile--language alignment is critical, particularly for descriptor generation. Overall, the ablation study shows that the gains of VitaTouch arise from the combination of multimodal fusion and language-anchored cross-modal alignment.

\begin{table}[!t]
    \centering
    \scriptsize
    \setlength{\tabcolsep}{4pt}
    \renewcommand{\arraystretch}{1.18}
    \caption{Comparison of the full VitaTouch model, a no-Stage-1 variant, and unimodal settings on VitaSet across tasks.}
    \label{tab:selfbuilt_summary}
    \begin{tabular}{@{}C{2.2cm}C{1.20cm}C{1.20cm}C{1.40cm}C{1.40cm}@{}}
        \toprule
        \textbf{Method} &
        \makecell[c]{\textbf{Hardness}\\\textbf{Acc.}} &
        \makecell[c]{\textbf{Roughness}\\\textbf{Acc.}} &
        \makecell[c]{\textbf{Descriptor}\\\textbf{Recall}} &
        \makecell[c]{\textbf{Mean Task}\\\textbf{Score}} \\
        \midrule
        \makecell[c]{Vision only\\(Qwen-2.5-VL)}
        & 73.81\% & 61.11\% & 32.28\% & 55.73\% \\
        \makecell[c]{Tactile only\\(Octopi)}
        & 68.25\% & 58.20\% & 8.40\% & 44.95\% \\
        \makecell[c]{VitaTouch\\(Without Stage 1)}
        & 84.80\% & 72.50\% & 17.46\% & 58.25\% \\
        \makecell[c]{\textbf{VitaTouch}\\\textbf{(Ours)}}
        & \textbf{88.89\%} & \textbf{75.13\%} & \textbf{54.81\%} & \textbf{72.95\%} \\
        \bottomrule
    \end{tabular}
\end{table}

\subsubsection{LoRA-Based Defect Adaptation for Data-Limited Manufacturing Inspection}
\label{subsubsec:stage3_fewshot}

We further evaluate VitaTouch on a separately collected defect-recognition dataset acquired under the same robotic inspection setup. This dataset is used only for Stage~3 adaptation and evaluation. We consider progressively finer 2-, 3-, and 5-category settings, and report few-shot results with $K \in \{5,15,50\}$ labeled training samples per category.

During Stage~3, the perception encoders and Q-Formers are frozen, and only the LoRA parameters in the LLM decoder are updated. Table~\ref{tab:defect_adaptation_comparison} shows that VitaTouch consistently outperforms the vision-only and tactile-only variants across all category settings and all values of $K$.

The advantage of multimodal adaptation remains evident under finer-grained label spaces. In the most challenging 5-category setting, VitaTouch reaches 85.3\% accuracy at $K=15$ and 92.0\% at $K=50$, outperforming the vision-only baseline (73.3\% and 86.0\%) and the tactile-only baseline (70.7\% and 82.0\%), respectively. These results show that LoRA-based adaptation can effectively exploit limited labeled data while preserving the benefit of multimodal grounding for fine-grained defect recognition.

\begin{table}[t]
\centering
\footnotesize
\setlength{\tabcolsep}{4.5pt}
\renewcommand{\arraystretch}{1.08}
\caption{LoRA-based defect adaptation results under different numbers of defect categories and labeled training samples per category. The vision-only and tactile-only baselines are built on Qwen-2.5-VL and Octopi, respectively.}
\label{tab:defect_adaptation_comparison}
\begin{tabular*}{\columnwidth}{@{\extracolsep{\fill}}ccccc@{}}
\toprule
\makecell[c]{Defect\\Categories} &
\makecell[c]{Train Samples\\per Category} &
\makecell[c]{Vision-only\\(Qwen-2.5-VL)} &
\makecell[c]{Tactile-only\\(Octopi)} &
\makecell[c]{VitaTouch\\(V+T)} \\
\midrule
2 & 5  & 70.0\% & 60.0\% & 90.0\% \\
2 & 15 & 76.7\% & 66.7\% & 96.7\% \\
2 & 50 & 82.0\% & 77.0\% & 100.0\% \\
\cmidrule(lr){1-5}
3 & 5  & 66.7\% & 66.7\% & 85.0\% \\
3 & 15 & 75.6\% & 73.3\% & 88.9\% \\
3 & 50 & 86.7\% & 81.3\% & 96.0\% \\
\cmidrule(lr){1-5}
5 & 5  & 64.0\% & 56.0\% & 72.0\% \\
5 & 15 & 73.3\% & 70.7\% & 85.3\% \\
5 & 50 & 86.0\% & 82.0\% & 92.0\% \\
\bottomrule
\end{tabular*}
\end{table}

\subsubsection{Quantitative Closed-Loop Robotic Sorting Validation}
\label{subsubsec:realworld_demo}

To assess practical deployability, we integrate VitaTouch into a laboratory robotic inspection and sorting system consisting of a Franka manipulator, a GelSight Mini tactile sensor mounted on the end effector, an external RGB camera, and a GPU workstation for inference. In each trial, the robot grasps the target object and acquires visual and tactile observations during interaction. These multimodal inputs are processed by the Stage~3 LoRA-adapted VitaTouch model to infer the defect status, which is subsequently translated into a binary task-level sorting command corresponding to either normal sorting or defect sorting. The corresponding motion planning and manipulation policy is then invoked to place the object into either the normal container or the defect container.

We evaluate the system using closed-loop recognition accuracy, end-to-end sorting success rate, and single-GPU inference latency. As summarized in Table~\ref{tab:robot_sorting_validation}, VitaTouch achieves 94.0\% closed-loop recognition accuracy and 94.0\% end-to-end sorting success rate over 100 trials, with a median latency of 278~ms per sample and a mean latency of 633~ms per sample. These results demonstrate the feasibility of deploying VitaTouch in a real robotic perception--decision--sorting loop under laboratory inspection conditions.

\begin{table}[t]
\centering
\caption{Quantitative results of the real-robot closed-loop sorting validation.}
\label{tab:robot_sorting_validation}
\begin{tabular}{l c}
\toprule
Metric & Result \\
\midrule
Number of closed-loop trials & 100 \\
Closed-loop recognition accuracy & 94.0\% \\
End-to-end sorting success rate & 94.0\% \\
Single-GPU inference latency (p50) & 278 ms/sample \\
Single-GPU inference latency (mean) & 633 ms/sample \\
\bottomrule
\end{tabular}
\end{table}

\begin{figure}[!t]
  \centering
  \includegraphics[width=\columnwidth]{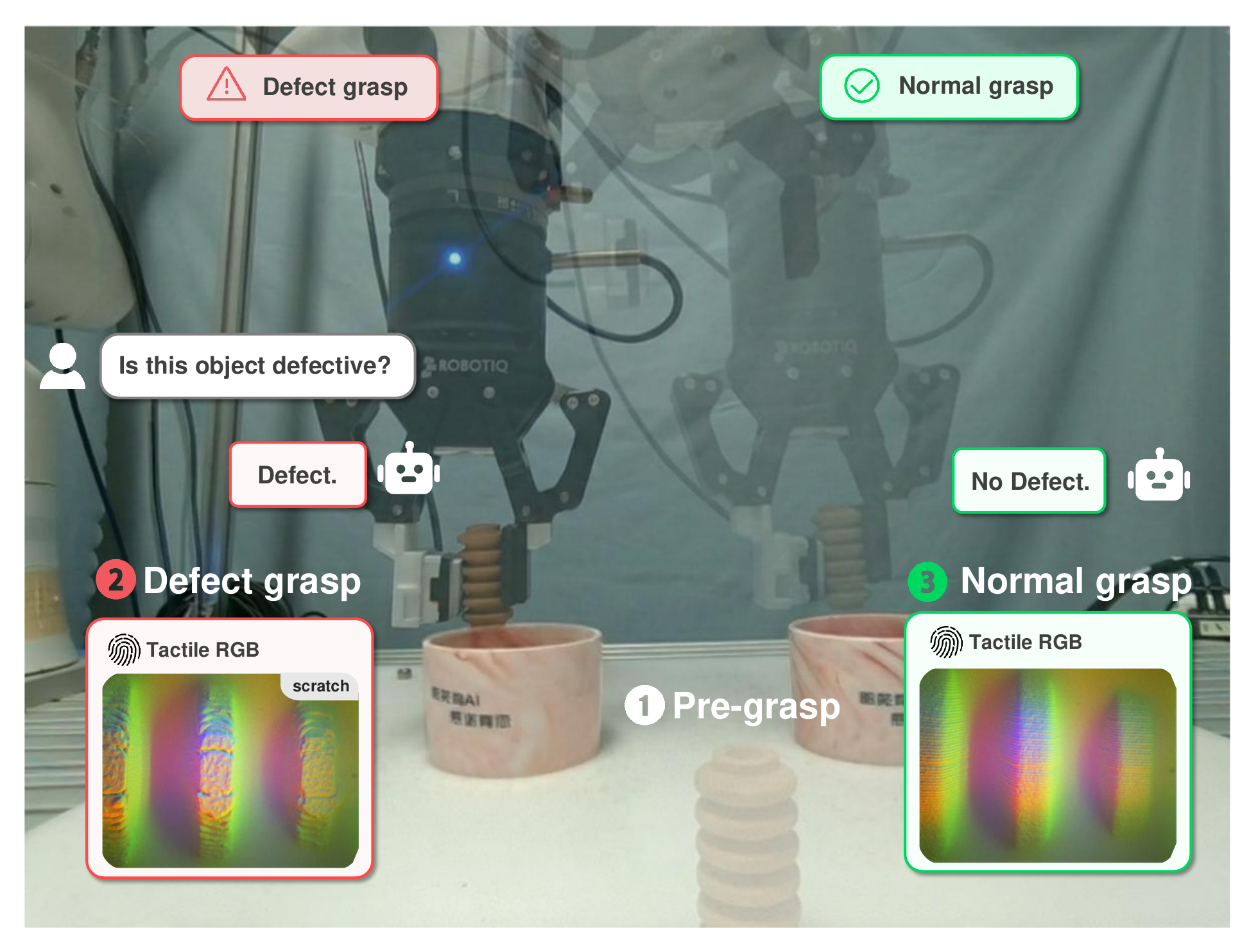}
  \caption{Closed-loop robotic inspection and sorting demonstration. A Franka manipulator grasps the object, acquires observations from an external RGB camera and the GelSight Mini tactile sensor during interaction, performs defect-status inference using VitaTouch, and places the object into the corresponding defect or normal container.}
  \label{fig:realworld_sorting}
\end{figure}

\section{Conclusion}
\label{sec:conclusion}

We presented \textbf{VitaTouch}, a property-aware vision--tactile--language model for manufacturing quality inspection. By combining dual Q-Formers with a frozen large language model, VitaTouch enables language-guided property understanding and defect reasoning from visual and tactile evidence.

Experiments demonstrate its effectiveness and transferability. On the public TVL benchmark, VitaTouch achieves the best performance on HCT and TVL among the compared methods. On VitaSet, it attains 88.89\% hardness accuracy, 75.13\% roughness accuracy, and 54.81\% descriptor recall, while the supplementary semantic-similarity score of 0.9009 further confirms the semantic consistency of the generated property descriptions. Under LoRA-based defect adaptation, VitaTouch also maintains strong performance on fine-grained defect classification, achieving 100.0\%, 96.0\%, and 92.0\% accuracy in 2-, 3-, and 5-category settings, respectively, with 50 labeled samples per category.

Beyond offline evaluation, we further validate VitaTouch in a closed-loop robotic sorting experiment, where the model is integrated into a perception-decision-sorting pipeline to support robotic defect sorting. These results indicate that property-oriented multimodal grounding is a promising direction for data-limited industrial inspection.

Nevertheless, the current study remains limited in experimental scale and environmental diversity. Future work will focus on larger-scale industrial validation, broader defect scenarios, and stronger robustness for real-world deployment.

\section*{Acknowledgment}
This work was supported by the Fundamental Research Funds for the Central Universities (Grant No. 2024PTB-007); the Fundamental Research Funds for Beijing University of Posts and Telecommunications (Grant No. 2025TSQY12); and Zhongguancun Academy (Grant No. 20240307).

\begin{IEEEbiographynophoto}{Junyi Zong}
received the B.Eng. degree in Energy and Power Engineering from Qingdao University of Science and Technology, Qingdao, China, in 2023. He is currently pursuing the Ph.D. degree in Control Science and Engineering at Beijing University of Posts and Telecommunications, Beijing, China. His research interests include multimodal large models, embodied intelligence, and robotic dexterous manipulation.
\end{IEEEbiographynophoto}

\begin{IEEEbiographynophoto}{Qingxuan Jia}
received the B.S. degree in mechanical design and manufacturing from Shandong University of Technology, China, in 1986, the M.S. degree in mechanics in 1991, and the Ph.D. degree in mechanical design and theory in 2005 from Beihang University, China. He is currently a Professor with the School of Intelligent Engineering and Automation, Beijing University of Posts and Telecommunications, China. His research interests include intelligent robotic technology, robotic manipulation, and system integration.
\end{IEEEbiographynophoto}

\begin{IEEEbiographynophoto}{Meixian Shi}
received the B.Eng. degree in Internet of Things Engineering from Hebei University of Technology, Tianjin, China, in 2025. She is currently pursuing the M.S. degree at the School of Intelligent Engineering and Automation, Beijing University of Posts and Telecommunications, Beijing, China. Her research interests include vision-language-action models, embodied control, and robotic manipulation.
\end{IEEEbiographynophoto}

\begin{IEEEbiographynophoto}{Tong Li}
received the B.S. degree in engineering mechanics from Beihang University (BUAA), Beijing, China, in 2010, and the Ph.D. degree in mechatronic engineering from Beijing University of Posts and Telecommunications (BUPT), Beijing, China, in 2016. He then worked as a Research Associate with the Aerospace Information Research Institute (AIR), Chinese Academy of Sciences, Beijing, China, until 2020. He is currently an Associate Professor with the School of Intelligent Engineering and Automation, BUPT. His research interests include tactile sensors, intelligent robotic technology, and multi-sensor information fusion.
\end{IEEEbiographynophoto}

\begin{IEEEbiographynophoto}{Jiayuan Li}
received the B.S. degree in control science and engineering from Beijing Institute of Technology, Beijing, China, in 2021, and the M.Eng. degree in computer control and automation from Nanyang Technological University, Singapore, in 2024. He is currently working toward the Ph.D. degree in control science and engineering at Beijing Institute of Technology, Beijing, China. His research interests include large language models, planning, combinatorial optimization, and continual learning.
\end{IEEEbiographynophoto}

\begin{IEEEbiographynophoto}{Zihang Lv}
received the B.Eng. degree in Automation from the School of Artificial Intelligence, Beijing University of Posts and Telecommunications, Beijing, China, in 2024. He is currently pursuing the M.S. degree in Control Science and Engineering at Beijing University of Posts and Telecommunications, Beijing, China. His research interest is robotic manipulation.
\end{IEEEbiographynophoto}

\begin{IEEEbiographynophoto}{Gang Chen}
received the B.S. degree in mechanical design, manufacture and automation from Beijing Institute of Petrochemical Technology (BIPT), Beijing, China, in 2004 and the Ph.D. degree in mechatronic engineering from Beijing University of Posts and Telecommunications (BUPT), Beijing, China, in 2011. He is currently a Professor with the School of Intelligent Engineering and Automation, BUPT. His research interests include space robotics, motion planning, and control methods.
\end{IEEEbiographynophoto}

\begin{IEEEbiographynophoto}{Fang Deng}
(Fellow, IEEE) received the B.Eng. and Ph.D. degrees in control science and engineering from Beijing Institute of Technology, Beijing, China, in 2004 and 2009, respectively. He is currently a Professor with the School of AI, Beijing Institute of Technology, Beijing, China, and also with Beijing Institute of Technology, Zhuhai, China. He serves as the Executive Vice President of both the Graduate School and the School of AI of Beijing Institute of Technology, and as the Director of the Beijing Key Laboratory of Lightweight Intelligent System. His research interests include swarm intelligence, autonomous sensor networks, and smart wearable devices.
\end{IEEEbiographynophoto}


\begin{thebibliography}{99}

\bibitem{azamfirei2023automation}
V.~Azamfirei, F.~Psarommatis, and Y.~Lagrosen, ``Application of automation for in-line quality inspection, a zero-defect manufacturing approach,'' \textit{J. Manuf. Syst.}, vol.~67, pp.~1--22, 2023.

\bibitem{lu2023deeplearning}
L.~Lu, J.~Hou, S.~Yuan, X.~Yao, Y.~Li, and J.~Zhu, ``Deep learning-assisted real-time defect detection and closed-loop adjustment for additive manufacturing of continuous fiber-reinforced polymer composites,'' \textit{Robot. Comput.-Integr. Manuf.}, vol.~79, Art. no.~102431, 2023.

\bibitem{reis2024inspection}
A.~M.~Reis, A.~Dall-Orsoletta, E.~Nunes, L.~Costa, and S.~Sousa, ``Quality costs and Industry 4.0: inspection strategy modelling and reviewing,'' \textit{Int. J. Adv. Manuf. Technol.}, vol.~136, no.~9, pp.~3883--3897, 2025.

\bibitem{liu2024iad_survey}
J.~Liu, G.~Xie, J.~Wang, S.~Li, C.~Wang, F.~Zheng, and Y.~Jin, ``Deep industrial image anomaly detection: A survey,'' \textit{Mach. Intell. Res.}, vol.~21, no.~1, pp.~104--135, 2024.

\bibitem{bergmann2019mvtec}
P.~Bergmann, M.~Fauser, D.~Sattlegger, and C.~Steger, ``MVTec AD---A comprehensive real-world dataset for unsupervised anomaly detection,'' in \textit{Proc. IEEE/CVF Conf. Comput. Vis. Pattern Recognit. (CVPR)}, 2019, pp.~9592--9600.

\bibitem{ma2024surface_review}
Y.~Ma, J.~Yin, F.~Huang, and Q.~Li, ``Surface defect inspection of industrial products with object detection deep networks: A systematic review,'' \textit{Artif. Intell. Rev.}, vol.~57, no.~12, Art. no.~333, 2024.

\bibitem{li2024tactilegrasping}
T.~Li, Y.~Yan, C.~Yu, J.~An, Y.~Wang, and G.~Chen, ``A comprehensive review of robot intelligent grasping based on tactile perception,'' \textit{Robot. Comput.-Integr. Manuf.}, vol.~90, Art. no.~102792, 2024.

\bibitem{xin2025vbtsreview}
Y.-H.~Xin, K.-M.~Hu, R.-J.~Xiang, Y.-L.~Gao, J.-F.~Zhou, G.~Meng, and W.-M.~Zhang, ``Vision-based tactile sensing: From performance parameters to device design,'' \textit{Appl. Phys. Rev.}, vol.~12, no.~2, Art. no.~021312, 2025.

\bibitem{feng2025anytouch}
R.~Feng, J.~Hu, W.~Xia, T.~Gao, A.~Shen, Y.~Sun, B.~Fang, and D.~Hu, ``AnyTouch: Learning unified static-dynamic representation across multiple visuo-tactile sensors,'' in \textit{Proc. Int. Conf. Learn. Represent. (ICLR)}, 2025.

\bibitem{radford2021clip}
A.~Radford, J.~W.~Kim, C.~Hallacy, A.~Ramesh, G.~Goh, S.~Agarwal, G.~Sastry, A.~Askell, P.~Mishkin, J.~Clark, G.~Krueger, and I.~Sutskever, ``Learning transferable visual models from natural language supervision,'' in \textit{Proc. 38th Int. Conf. Mach. Learn. (ICML)}, 2021, pp.~8748--8763.

\bibitem{li2023blip2}
J.~Li, D.~Li, S.~Savarese, and S.~Hoi, ``BLIP-2: Bootstrapping language-image pre-training with frozen image encoders and large language models,'' in \textit{Proc. 40th Int. Conf. Mach. Learn. (ICML)}, 2023, pp.~19730--19742.

\bibitem{dai2023instructblip}
W.~Dai, J.~Li, D.~Li, A.~M.~H.~Tiong, J.~Zhao, W.~Wang, B.~Li, P.~Fung, and S.~Hoi, ``InstructBLIP: Towards general-purpose vision-language models with instruction tuning,'' in \textit{Adv. Neural Inf. Process. Syst.}, vol.~36, 2023, pp.~49250--49267.

\bibitem{liu2023llava}
H.~Liu, C.~Li, Q.~Wu, and Y.~J.~Lee, ``Visual instruction tuning,'' in \textit{Adv. Neural Inf. Process. Syst.}, vol.~36, 2023, pp.~34892--34916.

\bibitem{fu2024tvl}
L.~Fu, G.~Datta, H.~Huang, W.~C.-H.~Panitch, J.~Drake, J.~Ortiz, M.~Mukadam, M.~Lambeta, R.~Calandra, and K.~Goldberg, ``A touch, vision, and language dataset for multimodal alignment,'' in \textit{Proc. 41st Int. Conf. Mach. Learn. (ICML)}, 2024, pp.~14080--14101.

\bibitem{yu2024octopi}
S.~Yu, K.~Lin, A.~Xiao, J.~Duan, and H.~Soh, ``Octopi: Object property reasoning with large tactile-language models,'' in \textit{Robotics: Science and Systems (RSS)}, 2024.

\bibitem{jiang2024mmad}
X.~Jiang, J.~Li, H.~Deng, Y.~Liu, B.-B.~Gao, Y.~Zhou, J.~Li, C.~Wang, and F.~Zheng, ``MMAD: A comprehensive benchmark for multimodal large language models in industrial anomaly detection,'' in \textit{Proc. Int. Conf. Learn. Represent. (ICLR)}, 2025.

\bibitem{defard2021padim}
T.~Defard, A.~Setkov, A.~Loesch, and R.~Audigier, ``PaDiM: A patch distribution modeling framework for anomaly detection and localization,'' in \textit{Comput. Anal. Images Patterns}, 2021, pp.~475--489.

\bibitem{zavrtanik2021draem}
V.~Zavrtanik, M.~Kristan, and D.~Sko{\v{c}}aj, ``DRAEM---A discriminatively trained reconstruction embedding for surface anomaly detection,'' in \textit{Proc. IEEE/CVF Int. Conf. Comput. Vis. (ICCV)}, 2021, pp.~8330--8339.

\bibitem{roth2022patchcore}
K.~Roth, L.~Pemula, J.~Zepeda, B.~Sch{\"o}lkopf, T.~Brox, and P.~Gehler, ``Towards total recall in industrial anomaly detection,'' in \textit{Proc. IEEE/CVF Conf. Comput. Vis. Pattern Recognit. (CVPR)}, 2022, pp.~14318--14328.

\bibitem{zou2022spd}
Y.~Zou, J.~Jeong, L.~Pemula, D.~Zhang, and O.~Dabeer, ``SPot-the-Difference self-supervised pre-training for anomaly detection and segmentation,'' in \textit{Comput. Vis.--ECCV 2022}, 2022, pp.~392--408.

\bibitem{zhou2025mvagfl}
J.~Zhou, M.~Liu, Y.~Ma, S.~Jiang, and Y.~Wang, ``Multi-view attention guided feature learning for unsupervised surface defect detection,'' \textit{IEEE/ASME Trans. Mechatronics}, vol.~30, no.~4, pp.~2844--2852, Aug.~2025, doi: 10.1109/TMECH.2025.3566311.

\bibitem{jeong2023winclip}
J.~Jeong, Y.~Zou, T.~Kim, D.~Zhang, A.~Ravichandran, and O.~Dabeer, ``WinCLIP: Zero-/few-shot anomaly classification and segmentation,'' in \textit{Proc. IEEE/CVF Conf. Comput. Vis. Pattern Recognit. (CVPR)}, 2023, pp.~19606--19616.

\bibitem{li2024promptad}
X.~Li, Z.~Zhang, X.~Tan, C.~Chen, Y.~Qu, Y.~Xie, and L.~Ma, ``PromptAD: Learning prompts with only normal samples for few-shot anomaly detection,'' in \textit{Proc. IEEE/CVF Conf. Comput. Vis. Pattern Recognit. (CVPR)}, 2024, pp.~16838--16848.

\bibitem{zhou2024anomalyclip}
Q.~Zhou, G.~Pang, Y.~Tian, S.~He, and J.~Chen, ``AnomalyCLIP: Object-agnostic prompt learning for zero-shot anomaly detection,'' in \textit{Proc. Int. Conf. Learn. Represent. (ICLR)}, 2024.

\bibitem{qu2024vcpclip}
Z.~Qu, X.~Tao, M.~Prasad, F.~Shen, Z.~Zhang, X.~Gong, and G.~Ding, ``VCP-CLIP: A visual context prompting model for zero-shot anomaly segmentation,'' in \textit{Comput. Vis.--ECCV 2024}, 2024, pp.~301--317.

\bibitem{jiang2025resilient}
S.~Jiang, Y.~Ma, J.~Zhou, Y.~Bian, Y.~Wang, and M.~Liu, ``Resilient multimodal industrial surface defect detection with uncertain sensors availability,'' \textit{IEEE/ASME Trans. Mechatronics}, vol.~30, no.~6, pp.~4261--4271, Dec.~2025, doi: 10.1109/TMECH.2025.3607147.

\bibitem{li2024clipsam}
S.~Li, J.~Cao, P.~Ye, Y.~Ding, C.~Tu, and T.~Chen, ``ClipSAM: CLIP and SAM collaboration for zero-shot anomaly segmentation,'' \textit{Neurocomputing}, vol.~618, Art. no.~129122, 2025.

\bibitem{ma2025aaclip}
W.~Ma, X.~Zhang, Q.~Yao, F.~Tang, C.~Wu, Y.~Li, R.~Yan, Z.~Jiang, and S.~K.~Zhou, ``AA-CLIP: Enhancing zero-shot anomaly detection via anomaly-aware CLIP,'' in \textit{Proc. IEEE/CVF Conf. Comput. Vis. Pattern Recognit. (CVPR)}, 2025, pp.~4744--4754.

\bibitem{gong2025feclip}
T.~Gong, Q.~Chu, B.~Liu, W.~Zhou, and N.~Yu, ``FE-CLIP: Frequency enhanced CLIP model for zero-shot anomaly detection and segmentation,'' in \textit{Proc. IEEE/CVF Int. Conf. Comput. Vis. (ICCV)}, 2025, pp.~21220--21230.

\bibitem{gu2023anomalygpt}
Z.~Gu, B.~Zhu, G.~Zhu, Y.~Chen, M.~Tang, and J.~Wang, ``AnomalyGPT: Detecting industrial anomalies using large vision-language models,'' in \textit{Proc. AAAI Conf. Artif. Intell.}, vol.~38, 2024, pp.~1932--1940.

\bibitem{yuan2017gelsight}
W.~Yuan, S.~Dong, and E.~H.~Adelson, ``GelSight: High-resolution robot tactile sensors for estimating geometry and force,'' \textit{Sensors}, vol.~17, no.~12, Art. no.~2762, 2017.

\bibitem{lambeta2020digit}
M.~Lambeta, P.-W.~Chou, S.~Tian, B.~Yang, B.~Maloon, V.~R.~Most, D.~Stroud, R.~Santos, A.~Byagowi, D.~Jayaraman, and R.~Calandra, ``DIGIT: A novel design for a low-cost compact high-resolution tactile sensor with application to in-hand manipulation,'' \textit{IEEE Robot. Autom. Lett.}, vol.~5, no.~3, pp.~3838--3845, 2020.

\bibitem{agarwal2023roboticdefect}
A.~Agarwal, A.~Ajith, C.~Wen, V.~Stryzheus, B.~Miller, M.~Chen, M.~K.~Johnson, J.~L.~Susa Rincon, J.~Rosca, and W.~Yuan, ``Robotic defect inspection with visual and tactile perception for large-scale components,'' in \textit{Proc. IEEE/RSJ Int. Conf. Intell. Robots Syst. (IROS)}, 2023, pp.~10110--10116.

\bibitem{finn2017maml}
C.~Finn, P.~Abbeel, and S.~Levine, ``Model-agnostic meta-learning for fast adaptation of deep networks,'' in \textit{Proc. 34th Int. Conf. Mach. Learn. (ICML)}, vol.~70, 2017, pp.~1126--1135.

\bibitem{snell2017protonet}
J.~Snell, K.~Swersky, and R.~Zemel, ``Prototypical networks for few-shot learning,'' in \textit{Adv. Neural Inf. Process. Syst.}, vol.~30, 2017.

\bibitem{hu2021lora}
E.~J.~Hu, Y.~Shen, P.~Wallis, Z.~Allen-Zhu, Y.~Li, S.~Wang, L.~Wang, and W.~Chen, ``LoRA: Low-rank adaptation of large language models,'' in \textit{Proc. Int. Conf. Learn. Represent. (ICLR)}, 2022.

\bibitem{zhang2025qwen3embedding}
Y.~Zhang, M.~Li, D.~Long, X.~Zhang, H.~Lin, B.~Yang, P.~Xie, A.~Yang, D.~Liu, J.~Lin, F.~Huang, and J.~Zhou,
``Qwen3 Embedding: Advancing text embedding and reranking through foundation models,''
\textit{arXiv preprint arXiv:2506.05176}, 2025.

\end{thebibliography}
\end{document}